\pdfoutput=1
\documentclass[10pt,twocolumn,letterpaper]{article}

\usepackage[pagenumbers]{wacv}   %

\usepackage{capt-of}
\usepackage{microtype}
\definecolor{wacvblue}{rgb}{0.21,0.49,0.74}
\usepackage[pagebackref,breaklinks,colorlinks,allcolors=wacvblue]{hyperref}
\hypersetup{pdftitle={Undertone: Personalizing Frozen VLM Aesthetic Judges from Few Ratings},
  pdfauthor={Amit Jadhav, Shaurya Beriwala, Beomjin Kim},
  pdfsubject={Personalized image aesthetics assessment},
  pdfkeywords={image aesthetics assessment, personalized image aesthetics, vision-language models, Bayesian linear regression}}
\def\wacvPaperID{*****}
\def\confName{WACV}
\def\confYear{2027}

\newcommand{\method}{\textup{\textsc{Undertone}}}       %
\newcommand{\methodP}{\mbox{\textup{\textsc{Undertone-P}}}}   %

\newcommand{\NdropIdfZS}{.845}

\newcommand{\NoffZSpanel}{.835}
\newcommand{\NaadbZSpanel}{.608}
\newcommand{\NoffBestPub}{.694}
\newcommand{\NaadbBestPub}{.514}
\newcommand{\NfactExpArgmax}{12 of 12}

\newcommand{\NjudgesBeatPub}{five}

\newcommand{\NpersRoneBten}{.641}
\newcommand{\NpersRoneBhund}{.667}

\newcommand{\NpersTrainB}{219}

\newcommand{\NpersRankMax}{218}

\newcommand{\NpersAblScoresHund}{.625}

\newcommand{\NpersAblJudgesHund}{.666}

\newcommand{\NdisjPhotos}{11,350}

\newcommand{\NhardOverlap}{31}

\newcommand{\NcostHol}{2.1}
\newcommand{\NvalidMassMin}{99.9\%}

\newcommand{\NpanelInt}{.822}
\newcommand{\NcontamIdf}{.039}
\newcommand{\NcontamOthersMax}{.012}

\newcommand{\NholBten}{.648}
\newcommand{\NholBhund}{.676}
\newcommand{\NholBtenCI}{[.629, .667]}
\newcommand{\NholBhundCI}{[.660, .693]}
\newcommand{\NholPopB}{.640}

\newcommand{\NholDims}{390}
\newcommand{\NholDimsAll}{391}

\newcommand{\NholAten}{.659}
\newcommand{\NholAhund}{.678}

\newcommand{\NholApopTen}{.652}
\newcommand{\NholAaddTen}{.007}
\newcommand{\NholAaddHund}{.029}

\newcommand{\NholHardHund}{.524}

\newcommand{\NlapTen}{.532}
\newcommand{\NlapHund}{.627}

\newcommand{\NlapPop}{.421}

\newcommand{\NlapTrain}{367}

\newcommand{\NlapRaters}{567}

\newcommand{\NonlyRatLapTen}{.265}
\newcommand{\NonlyRatLapHund}{.554}

\newcommand{\NoneJudgeTenMin}{.616}
\newcommand{\NoneJudgeTenMax}{.641}
\newcommand{\NoneJudgeHundMin}{.641}
\newcommand{\NoneJudgeHundMax}{.668}
\newcommand{\NholAddTenPt}{$+$.008}
\newcommand{\NholAddHundPt}{$+$.037}
\newcommand{\NholHardAddHundPt}{$+$.067}
\newcommand{\NlapAddTenPt}{$+$.111}
\newcommand{\NlapAddHundPt}{$+$.206}

\newcommand{\NholCovIsoHundPt}{$+$.021}

\newcommand{\NholCovDiagHundPt}{$+$.012}
\newcommand{\NholOwnVsOtherHundPt}{$+$.051}
\newcommand{\NholOwnVsCrowdHundPt}{$+$.038}

\newcommand{\NpersAblHiddenGainPt}{$+$.041}

\newcommand{\NcrIndFullTen}{.645}
\newcommand{\NcrIndFullHund}{.671}

\newcommand{\NcrCleanTenMin}{.626}
\newcommand{\NcrCleanTenMax}{.639}
\newcommand{\NcrCleanHundMin}{.651}
\newcommand{\NcrCleanHundMax}{.663}

\newcommand{\NcrLapOneHundMin}{.574}
\newcommand{\NcrLapOneHundMax}{.605}

\newcommand{\NcrXferTen}{.430}
\newcommand{\NcrXferHund}{.573}

\newcommand{\NcrXferGapTen}{.102}
\newcommand{\NcrXferGapHund}{.054}

\newcommand{\NcrLeadThou}{$+$.008}

\newcommand{\NcrThouRaters}{140}

\newcommand{\NcrHundToFive}{$+$.017}
\newcommand{\NcrThouMinusFive}{$+$.006}

\newcommand{\NcrFewMinusPriorMax}{$+$.003}
\newcommand{\NcrUserMs}{1.2}

\title{Undertone: Personalizing Frozen VLM Aesthetic Judges from Few Ratings}

\author{Amit Jadhav$^*$ \qquad Shaurya Beriwala$^*$ \qquad Beomjin Kim\\
Purdue University Fort Wayne\\
{\tt\small \{jadhah01, beris01, kim43\}@purdue.edu} \qquad {\small $^*$Equal contribution}}

\begin{document}
\AddToShipoutPicture*{\AtPageLowerLeft{\put(0.8125in,0.22in){\parbox[b]{6.875in}{\centering\fontsize{6.5}{7.6}\selectfont\color[gray]{0.35}Accepted at the WVAQ Workshop at WACV 2027 (IEEE/CVF WACV Workshops). \copyright\ 2027 IEEE. Personal use of this material is permitted. Permission from IEEE must be obtained for all other uses, in any current or future media, including reprinting/republishing this material for advertising or promotional purposes, creating new collective works, for resale or redistribution to servers or lists, or reuse of any copyrighted component of this work in other works.}}}}

\twocolumn[{%
\renewcommand\twocolumn[1][]{#1}%
\maketitle
\vspace{-2.2em}
\begin{center}
  \includegraphics[width=\textwidth]{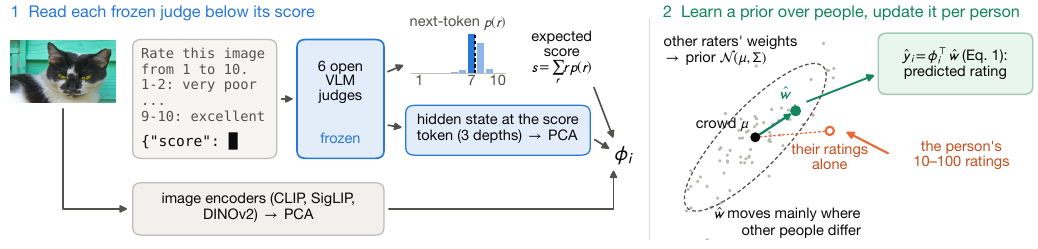}
  \vspace{-1.5em}
  \captionof{figure}{\textbf{\method{} at a glance.} (1)~\emph{Read-outs}: six frozen vision-language model (VLM)
  judges rate the image; we read the expected score $s$ (dashed: probability-weighted, not the integer) and
  the hidden state at the score token, plus three frozen image encoders, together the read-out vector $\phi_i$.
  (2)~\emph{Personalization}: other raters' regression weights on $\phi$ (grey) give a Gaussian prior around their mean
  $\mu$, the crowd model; a person's ratings move the estimate $\hat w$ toward their own fit (orange), mostly along
  directions in which people differ (Eq.~\ref{eq:person}; a computed 2-D example).}
  \label{fig:overview}
\end{center}
\vspace{0.4em}
}]

\begin{abstract}
Image quality and aesthetics models predict the average rating of many people, but tastes differ, so a personalized
model must learn each new user's taste from the few images they rate. Fitting a model to one person from scratch fails
with ten ratings, and networks fine-tuned per person carry over poorly to new kinds of images. We present \methodP,
which trains no network. Frozen vision-language model (VLM) judges describe each image by the probability they give each
rating and by their internal state. From many other raters, \methodP{} learns a Gaussian prior over how a
person's taste departs from the crowd's; a new user's few ratings then only choose among these known departures, in one
closed-form update. On Ryu and Yanaka's PARA benchmark, the prior alone ranks a person's unseen photographs with a
Spearman correlation of \NholPopB, above the best published results with 10 and 100 ratings (.581 and .620); 10 and 100
of the person's own ratings raise it to \NholBten{} and \NholBhund. These ratings help most where tastes depart from the
crowd: with 100 ratings, \NholHardAddHundPt{} for the raters who agree least with it and \NlapAddHundPt{} on LAPIS
artworks, against \NholAddHundPt{} overall. Without any ratings, the judges predict PARA's average scores at
\NoffZSpanel, above every published model we ran. Networks trained on PARA still lead under PARA's
original protocol.
\end{abstract}

\section{Introduction}

Tastes differ: the same photograph can delight one viewer and bore another. Image quality assessment predicts the mean
opinion score (MOS) of many raters, but for aesthetics the mean hides the individual, so an image platform should rank
images the way each of its users would. Personalized image aesthetics assessment (PIAA) predicts one person's ratings
from a few of them \cite{ren2017personalized,yang2022para}, and the cold start, a new user with only a handful of
ratings, matters most. The usual recipe pretrains an image network on tens of thousands of rated images and fine-tunes
it for each person \cite{yang2022para,zhu2022blgpiaa,shi2024piaaici,gao2024aesmamba}. Such networks carry over poorly
to new kinds of images: PIAA-ICI \cite{shi2024piaaici}, trained on the photographs of PARA
(Personalized image Aesthetics database with Rich Attributes) \cite{yang2022para}, reaches only .206 on artworks
\cite{ryu2026piaa}. Vision-language models (VLMs) offer a route without training: asked to rate an image, a VLM acts as
a \emph{judge}. Ryu and Yanaka \cite{ryu2026piaa} (R\&Y) fit a linear probe on a frozen VLM's hidden state for each
person. With 100 ratings it beats prompting and fine-tuning with low-rank adaptation (LoRA), but with 10 it reaches only
.389--.402, below simply asking the VLM (.570): ten ratings are too few to learn a taste from scratch.

Our key idea is a prior over people (Fig.~\ref{fig:overview}). People differ from the MOS in ways that other raters
reveal. A few ratings cannot build a model of a person's taste, but they can correct a prior over per-person weights
learned from many other people: the prediction starts from a MOS predictor and moves toward the person's own taste as
ratings accumulate. The prior needs rich frozen features, so we read VLM judges below their integer rating, as is
established for quality assessment and text judges \cite{liu2023geval,wu2024qalign,you2025deqa}: before answering ``7''
a judge assigns a probability to every rating, and its hidden state describes the image in far more detail.

\method{} reads frozen judges below their rating; \methodP{} (P for personalized) learns the prior on these read-outs and
updates it in closed form. It is built for the cold start: a new user rates about ten images and costs one linear solve (\NcrUserMs{} ms on one
laptop CPU core). No network is trained, and a new kind of image needs a new prior, not a new network. Scoring costs
\NcostHol{} A100 GPU-seconds per image, paid once; on photographs, one 7B judge alone also works. Our contributions:
\begin{itemize}\setlength\itemsep{1pt}
  \item \textbf{A cold-start method that trains no network:} a prior over per-person weights, learned from other raters,
    on frozen VLM read-outs (Sec.~\ref{sec:method}).
  \item \textbf{Above every published result from few ratings} on R\&Y's PARA benchmark (even with one judge), on
    its hardest raters, and on artworks (other raters); without labels, the judges alone are a strong MOS predictor (Sec.~\ref{sec:exp}).
  \item \textbf{Where personalization matters:} with few ratings the prior, a MOS predictor, carries most of the
    accuracy; a person's ratings add most where tastes depart from the crowd (hard raters, artworks). Plus why networks
    trained on PARA lead under PARA's protocol (Sec.~\ref{sec:person}).
\end{itemize}

\textbf{Related work.} Aesthetic models learn from large rated collections
\cite{talebi2018nima,ke2021musiq,wu2024qalign,gao2024aesmamba}, and VLM raters are trained for the task
\cite{wu2024qalign,humanAesExpert2025,liu2025aesr1,jia2026redaes} or prompted; HumanAesExpert \cite{humanAesExpert2025} trains heads on the expected score and the hidden
state. PIAA methods learn from other raters by training networks
\cite{yang2022para,zhu2022blgpiaa,shi2024piaaici,yun2024taskvector,yang2026prac}; \citet{ren2017personalized} fit a per-user residual
on a generic model; a linear version is our method with an isotropic prior (Sec.~\ref{sec:ablate}). Priors over per-user parameters learned from a population are classic in Bayesian statistics
\cite{lindley1972bayes,hachemeister1975credibility,allenby1998heterogeneity} and appear in multi-task learning
\cite{bakker2003taskclustering}, recommendation \cite{zhang2007hierarchical} and reward models
\cite{chen2024pal,bose2025lore}; we bring them to the read-outs of frozen VLM judges.

\section{Method}
\label{sec:method}

Fig.~\ref{fig:overview} shows the two stages: frozen models turn each image into read-outs, and a prior learned from
other raters turns a person's few ratings into predictions.

\textbf{Terms.} A \emph{judge} is a VLM prompted to rate an image from 1 to 10; its \emph{read-outs} are numbers we take
from it. The \emph{crowd} is all raters of a dataset, a \emph{crowd model} predicts their mean opinion score (MOS), and a
\emph{person} is one rater. We score predictions by Spearman's rank correlation $\rho$ between predicted and actual
ratings (1: same order; 0: unrelated); for a person, on held-out images, averaged over persons.

\subsection{Reading a frozen judge}
Each judge receives the image and a prompt asking for an overall score from 1 to 10 with five score bands (``1--2: very
poor \dots{} 9--10: excellent''). We pre-fill its answer with \texttt{\{"score":} and read the next-token probability
$p(r)$ of each rating $r=1,\dots,10$, renormalized to sum to one (they receive over \NvalidMassMin{} of the probability on
average).
The first read-out is the \emph{expected score} $s=\sum_r r\,p(r)$; the usual answer keeps only the most likely $r$. The
second is the \emph{hidden state} at the score token, the last input token, where the judge is about to write its rating, at half,
three-quarter and full depth. Principal component analysis (PCA) reduces it to 64 components per judge and depth, and
again to 64 per depth across the six judges. Three frozen image encoders, CLIP \cite{cherti2023openclip}, SigLIP
\cite{zhai2023siglip} and DINOv2 \cite{oquab2023dinov2}, add 64 principal components each. Every PCA is fitted on all
images, without ratings; fitting every PCA and the prior only on the \NdisjPhotos{} photographs outside every R\&Y
evaluation set changes \methodP's results in Sec.~\ref{sec:person} from \NholBten{} and \NholBhund{} to \NcrIndFullTen{}
and \NcrIndFullHund. The read-out vector $\phi_i$ of image $i$ holds an intercept and these
\NholDims{} standardized numbers. Without labels, the \emph{zero-shot panel} predicts the MOS as
$a_i=\frac1J\sum_{j}z(s_{ij})$, the mean of the $J=6$ judges' standardized expected scores.

\subsection{\methodP: a prior over people}
We model person $u$'s rating of image $i$ as $y_{ui}=\phi_i^\top w_u+\varepsilon$ with noise variance $\sigma^2$. Ten
ratings cannot determine the weights $w_u$, so we learn from other people what such weights look like. Each other rater
$v$ gets ridge-regression weights $\hat w_v$ (penalty 1000; on the R\&Y benchmark of Sec.~\ref{sec:exp}, all \NpersTrainB{} non-test raters). Their mean $\mu$ is a crowd model, their sample
covariance $V$ describes how people differ from it, and their mean residual variance gives $\sigma^2$. With the prior
$w_u\sim\mathcal N(\mu,\Sigma)$, $\Sigma=cV+10^{-6}I$ and $c=1$, the person's ratings $y_S$ of images with read-outs
$X_S$ give the posterior mean $\hat w$ of Bayesian linear regression \cite{lindley1972bayes} and the predictions
$\hat y_i=\phi_i^\top \hat w$:
\begin{equation}
  \hat w=\big(\Sigma^{-1}+X_S^\top X_S/\sigma^2\big)^{-1}\big(\Sigma^{-1}\mu+X_S^\top y_S/\sigma^2\big).
  \label{eq:person}
\end{equation}

\textbf{Why few ratings suffice.} With no ratings $\hat w=\mu$, the crowd model. Each rating pulls $\hat w$ toward the
person's own fit, but mostly along directions in which other people's weights vary: $V$ has rank at most
\NpersRankMax{} (raters minus one) in \NholDimsAll{} dimensions, and elsewhere the $10^{-6}$ term allows only tiny moves. A
few ratings therefore choose among the ways in which people are known to differ, not among all possible weights; as
ratings accumulate, the person's own fit takes over.

\textbf{No leakage.} Test raters' ratings never enter $\mu$, $\Sigma$ or $\sigma^2$. For evaluation we also \emph{cross-fit}: images are
split into five groups, and each group is predicted with a prior learned only from ratings of images in the other four
(deployment needs one prior). The read-outs, the ridge penalty and $c$ were chosen on two sets of 40
non-test validation raters.

\section{Experiments}
\label{sec:exp}

\textbf{Data.} PARA \cite{yang2022para}: 31,220 photographs, each rated 1--5 by about 25 of 438 raters; its official
3,000-photograph test split serves crowd scores, and all photographs and raters serve personalization. AADB (Aesthetics
and Attributes DataBase) \cite{kong2016photo}: its official 1,000-photograph test split (5 raters each), for crowd
scores. LAPIS (Leuven Art Personalized Image Set) \cite{maerten2025lapis}: 11,723 artworks, each rated 0--100 by about
24 of \NlapRaters{} raters (cleaned 2026 release), for personalization.

\textbf{Judges.} Six open-weight VLMs of 7--12 billion parameters, all frozen: Qwen2.5-VL-7B
\cite{bai2025qwen25vl}, Qwen3-VL-8B \cite{qwen2025qwen3vl}, InternVL3-8B \cite{zhu2025internvl3}, LLaVA-OneVision-7B
(LLaVA-OV-7B) \cite{li2024llavaonevision}, Idefics3-8B \cite{laurencon2024idefics3} and Pixtral-12B \cite{agrawal2024pixtral}.

\textbf{Baselines.} On crowd scores we ran NIMA \cite{talebi2018nima}, MUSIQ \cite{ke2021musiq}, CLIP-IQA+
\cite{clipiqa2023}, the LAION aesthetic predictor \cite{laionaesthetics2022} (with PyIQA \cite{chen2022pyiqa}) and
Q-Align \cite{wu2024qalign} (OneAlign) on the same test photographs. On personalization we compare with results
reported on the same benchmarks, and with our read-outs fitted to each person's ratings alone by ridge regression
(``person's ratings only''; penalty set on validation raters).

\textbf{Protocols for one person.} The \emph{R\&Y benchmark} \cite{ryu2026piaa} fixes 200 PARA raters, each with 10 or
100 ratings to learn from and 50 other photographs to test on; its 50 \emph{hard raters}, who agree least with the crowd,
form a second test group (\NhardOverlap{} also among the 200). PARA's original protocol \cite{yang2022para} (the
\emph{PARA protocol}) uses 40 raters with at least 150 ratings (ours: a random draw) and random
draws of 10 or 100 ratings and 50 test photographs (our prior learns from all other raters); we report the mean
$\pm$ standard deviation (s.d.)\ over 10 repeats of 10 draws per rater. On LAPIS we
apply R\&Y's split code to the cleaned release (one rater and duplicates removed), so our 200 raters differ from theirs.

\textbf{Fair comparison.} Every method may use the benchmark's non-test raters; the trained networks and \methodP{} do,
R\&Y's methods do not. An earlier version of our method, also tuned on validation raters, scored \NpersRoneBten{} and
\NpersRoneBhund{} on the R\&Y benchmark. Gains we report on the R\&Y benchmark and
LAPIS with 10+ ratings are significant: 95\% bootstrap intervals over raters exclude zero.

\begin{table}[t]
\centering\footnotesize
\setlength{\tabcolsep}{3pt}
\caption{\textbf{MOS prediction without labels.} Spearman $\rho$ with the crowd's mean rating; ours on the official test splits (PARA 3,000, AADB 1,000 photographs). No published model listed reports training on PARA or AADB. $^a$Lower, not shown: MUSIQ, CLIP-IQA+ (PARA $\le$.499, AADB $\le$.354). $^b$Qwen3-VL-235B, Qwen2.5-VL-72B, GPT-5. $^c$Trained on edit pairs scored by that vote. Bold: best per column.}
\label{tab:zs}
\begin{tabular}{@{}lcc@{}}
\toprule
 & PARA & AADB \\
\midrule
\multicolumn{3}{@{}l}{\emph{Published aesthetic models, run by us$^a$}} \\
\quad LAION aesthetic predictor \cite{laionaesthetics2022} & .620 & .402 \\
\quad NIMA \cite{talebi2018nima} & .623 & .429 \\
\quad Q-Align (OneAlign) \cite{wu2024qalign} & .694 & .514 \\
\multicolumn{3}{@{}l}{\emph{Six open judges (ours)}} \\
\quad One judge, expected score (range) & .683--.828 & .413--.599 \\
\quad Panel of the integer ratings & .822 & .594 \\
\quad \method{} (panel of expected scores) & .835 & .608 \\
\multicolumn{3}{@{}l}{\emph{Reported by others}} \\
\quad GPT-5 \cite{jia2026redaes} & .735 & .599 \\
\quad Vote of three larger VLMs$^b$ \cite{jia2026redaes} & .796 & .677 \\
\quad RED-Aes-7B, trained$^c$ \cite{jia2026redaes} & \textbf{.859} & \textbf{.775} \\
\bottomrule
\end{tabular}
\end{table}

\subsection{MOS prediction without labels}
\label{sec:zs}

On PARA (Table~\ref{tab:zs}) the zero-shot panel reaches \NoffZSpanel, against \NoffBestPub{} for the best published
model we ran (Q-Align), and \NjudgesBeatPub{} of the six judges beat every published model we ran on their own. The
expected score beats the integer answer for every judge (\NfactExpArgmax{} judge--dataset pairs), less so in the panel
(\NpanelInt{} with integers). Of the listed models, only RED-Aes, a VLM trained on edit pairs, scores higher on PARA; models
trained on PARA's own labels reach .91--.94 \cite{cao2025artimuse}. On AADB the panel leads the published models we ran
(\NaadbZSpanel{} vs.\ \NaadbBestPub), but others report higher results (up to .775 \cite{jia2026redaes}).

\begin{table*}[t]
\centering\footnotesize
\setlength{\tabcolsep}{3.6pt}\renewcommand{\arraystretch}{0.92}
\caption{\textbf{Personalization from 10 or 100 of a person's ratings.} Spearman $\rho$ on each person's held-out images, mean over persons (PARA protocol: $\pm$ s.d.\ over 10 repeats of 10 draws). $^\dag$Network trained on PARA; these and \methodP{} learn from other raters, R\&Y's methods and `person's ratings only' do not. $^a$R\&Y's best per column (LoRA at 10 on PARA, else linear probes). $^b$Trained by R\&Y. $^c$Its authors', quoted by \cite{maerten2025lapis}. $^d$Five more prompts per judge, six times the cost. LAPIS: our raters, not R\&Y's. $^e$Best judge and depth on the test raters. Text gains use unrounded values.}
\label{tab:pers}
\begin{tabular}{@{}lcccccccc@{}}
\toprule
 & \multicolumn{6}{c}{PARA (photographs)} & \multicolumn{2}{c}{LAPIS (artworks)} \\
\cmidrule(lr){2-7}\cmidrule(lr){8-9}
 & \multicolumn{2}{c}{R\&Y benchmark} & \multicolumn{2}{c}{hard raters} & \multicolumn{2}{c}{PARA protocol} & \multicolumn{2}{c}{R\&Y procedure} \\
\cmidrule(lr){2-3}\cmidrule(lr){4-5}\cmidrule(lr){6-7}\cmidrule(lr){8-9}
Ratings from the person & 10 & 100 & 10 & 100 & 10 & 100 & 10 & 100 \\
\midrule
\multicolumn{9}{@{}l}{\emph{Reported by others}} \\
\quad Best VLM method$^a$ \cite{ryu2026piaa} & .581 & .611 & -- & .472 & -- & -- & .409 & .595 \\
\quad PIAA-ICI$^\dag$ \cite{shi2024piaaici} & -- & .620$^b$ & -- & -- & -- & \textbf{.739}$^c$ & -- & -- \\
\quad AesMamba-P$^\dag$ \cite{gao2024aesmamba} & -- & -- & -- & -- & \textbf{.707} & .723 & -- & -- \\
\quad PRAC$^\dag$ \cite{yang2026prac} & -- & -- & -- & -- & \textbf{.707} & .733 & -- & -- \\
\quad ResNet-50$^\dag$ \cite{yang2022para} & -- & -- & -- & -- & .681 & .695 & -- & -- \\
\multicolumn{9}{@{}l}{\emph{Ours (no network trained)}} \\
\quad Zero-shot panel (no ratings) & .600 & .600 & .435 & .435 & .610\,$\pm$\,.004 & .607\,$\pm$\,.004 & .281 & .281 \\
\quad Person's ratings only (no prior) & .325 & .606 & .246 & .464 & -- & -- & .265 & .554 \\
\quad R\&Y's probe, rerun with our judges$^e$ & .486 & .632 & -- & -- & -- & -- & -- & -- \\
\quad Prior alone (no ratings from the person) & .640 & .640 & .457 & .457 & .652\,$\pm$\,.004 & .650\,$\pm$\,.004 & .421 & .421 \\
\quad \textbf{\methodP{}} & \textbf{.648} & \textbf{.676} & \textbf{.473} & \textbf{.524} & .659\,$\pm$\,.004 & .678\,$\pm$\,.003 & \textbf{.532} & \textbf{.627} \\
\quad \methodP{}, five-prompt variant$^d$ & .646 & .674 & .469 & .521 & .657\,$\pm$\,.003 & .677\,$\pm$\,.004 & -- & -- \\
\bottomrule
\end{tabular}
\end{table*}

\begin{figure}[t]
  \centering
  \includegraphics[width=\linewidth]{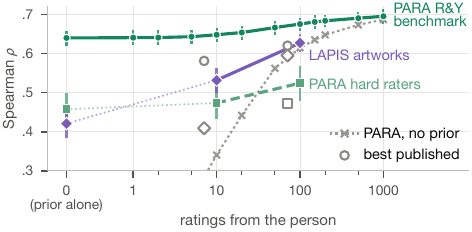}
  \caption{\textbf{Prior first, then the person.} \methodP's $\rho$ (95\% intervals) against the person's number of
  ratings. R\&Y benchmark and crosses: ratings drawn at random, unlike Table~\ref{tab:pers} (140 raters at
  1,000). Crosses: the person's ratings alone. Hollow: best published (LAPIS: other raters).}
  \label{fig:curves}
\end{figure}

\subsection{Personalization from few ratings}
\label{sec:person}

\textbf{Against published results.} On the R\&Y benchmark (Table~\ref{tab:pers}, Fig.~\ref{fig:curves}), \methodP{} reaches \NholBten{} \NholBtenCI{} with
10 of a person's ratings and \NholBhund{} \NholBhundCI{} with 100 (95\% intervals), above the best published
.581 and .620. Our read-outs fitted to the person alone, and R\&Y's probe rerun with our judges, stay below \methodP{}
(Table~\ref{tab:pers}): the gain comes from the prior, not from reading six judges. Even a single judge, with its own read-outs and no image encoders, reaches
\NoneJudgeTenMin--\NoneJudgeTenMax{} with 10 ratings and \NoneJudgeHundMin--\NoneJudgeHundMax{} with 100,
above every published result. The best published 100-rating result, .620, is PIAA-ICI as trained by R\&Y; its
authors report .739 under the PARA protocol (below), so this comparison may favor us. On the 50
hard raters \methodP{} reaches \NholHardHund{} with 100 ratings, against .472.

\textbf{Prior and person share the work.} On the R\&Y benchmark the prior alone reaches \NholPopB, already above every published result, because on PARA
most raters' tastes are close to the crowd's. The person's ratings add only \NholAddTenPt{} with 10 and \NholAddHundPt{} with 100,
and more where tastes depart from the crowd: \NholHardAddHundPt{} for the hard raters and \NlapAddHundPt{} on LAPIS.
They must be the person's own: at 100 ratings they beat random other raters' ratings by
\NholOwnVsOtherHundPt{} and the crowd mean by \NholOwnVsCrowdHundPt.

\textbf{From 1 to 1,000 ratings} (random draws; Fig.~\ref{fig:curves}). One to five ratings barely move \methodP{} from
the prior ($\le$\NcrFewMinusPriorMax); it leads the person's ratings alone at every budget. On the \NcrThouRaters{}
raters with 1,000 ratings, it gains \NcrHundToFive{} from 100 to 500 and \NcrThouMinusFive{} from 500 to 1,000, and its
lead shrinks to \NcrLeadThou: the prior helps most in the cold start.

\textbf{Artworks (LAPIS).} \methodP{} reaches \NlapTen{} with 10 ratings and \NlapHund{} with 100 (PARA's settings
unchanged, prior refit on LAPIS's \NlapTrain{} other raters). That is above the person's ratings alone on the same raters
(\NonlyRatLapTen{} and \NonlyRatLapHund) and the .409 and .595 that R\&Y report for other raters. Tastes depart further from the
crowd: the prior alone reaches \NlapPop, and the person's ratings add \NlapAddTenPt{} with 10 and
\NlapAddHundPt{} with 100.
One judge alone reaches \NcrLapOneHundMin{} (Idefics3-8B) to \NcrLapOneHundMax{} (LLaVA-OV-7B) with 100 (worst
and best, as on PARA). A prior and PCA from PARA give \NcrXferTen{} and \NcrXferHund, far above PIAA-ICI
trained on PARA (.206 at 100, other raters \cite{ryu2026piaa}) but \NcrXferGapTen{} and \NcrXferGapHund{} below
\methodP{} with LAPIS's prior.

\textbf{Why PARA-trained networks lead.} Under the PARA protocol \methodP{} reaches \NholAten{} and \NholAhund{}, below the PARA-trained networks (Table~\ref{tab:pers}). These networks
are trained end to end on PARA and start from a crowd model fitted to it: before personal fine-tuning the PARA
ResNet-50 reaches .6808 per rater \cite{yang2022para}, our prior \NholApopTen{} (different raters). The personal step adds more to \methodP{}
(\NholAaddTen{} and \NholAaddHund) than to that network (.0003 and .014). Their lead likely comes from end-to-end training,
costly to redo per domain.

\subsection{What each part contributes}
\label{sec:ablate}

\textbf{Prior} (100 ratings). The learned $\Sigma$ beats a diagonal one by \NholCovDiagHundPt{} and an isotropic one, i.e.\
a linear version of the residual fit of \cite{ren2017personalized}, by \NholCovIsoHundPt{} (set on
validation raters). \textbf{Five-prompt variant} (Table~\ref{tab:pers}): at 100 ratings and without image encoders, the judges' expected scores give \NpersAblScoresHund, and their hidden states add
\NpersAblHiddenGainPt{} to \NpersAblJudgesHund, the largest gain of any read-out.

\section{Limitations and conclusion}
\label{sec:end}

A prior over people, learned from other raters on frozen VLM read-outs, personalizes aesthetic scores from a few
ratings without training a network, above every published result on R\&Y's benchmark. The prior carries most of the
accuracy; a person's ratings matter most where tastes depart from the crowd. \textbf{Limitations.} Like trained networks, the prior needs many raters of similar
images; a stronger crowd model inside it could close the gap to PARA-trained networks.
Judges may have seen PARA in training: Idefics3-8B's $\rho$
is \NcontamIdf{} higher on PARA's training than test photographs (others $\le$~\NcontamOthersMax);
without it the panel reaches \NdropIdfZS. Qwen2.5-VL-7B and LLaVA-OV-7B were evaluated on PARA's test split in our earlier
work; Qwen3-VL-8B, InternVL3-8B and \mbox{Pixtral-12B} alone still reach \NcrCleanTenMin--\NcrCleanTenMax{} and
\NcrCleanHundMin--\NcrCleanHundMax{} with 10 and 100, above every published result.
\textbf{Broader impact.} The prior pulls predictions toward majority taste, so users unlike any rater are least
served; taste profiles are personal data.

{
    \small
    \bibliographystyle{ieeenat_fullname}
    \bibliography{references}
}

\end{document}